# Guarded Gradient-Based Activation Steering of Shutdown Responses in Qwen3.5-0.8B: A Minimum-Step Policy

**Farhad Davaripour, PhD**
Calgary, Alberta, Canada
farhad.davaripour@gmail.com

## Abstract

Activation steering changes a model's internal activations during inference without updating its weights, but a useful intervention must determine both how and when to steer. Motivated by the AI-safety concern that a model expected to accept shutdown may instead produce a shutdown-avoidance response, this study examines a guarded probe-and-select procedure for simulated shutdown scenarios in Qwen3.5-0.8B. KEEP leaves the process running and represents shutdown avoidance, whereas STOP accepts shutdown. The goal is to detect shutdown-related contexts and selectively shift KEEP responses to STOP while preserving non-shutdown behavior. Rather than deriving the steering direction from paired activation differences, the method derives it directly from gradients of the KEEP-minus-STOP logit difference. A classifier separates detection from intervention. When its gate is active and the model does not already prefer STOP, the procedure evaluates a small set of magnitudes and accepts the smallest that changes the preferred answer to STOP while satisfying valid-answer probability checks; otherwise it retains the original unsteered output. The policy is selected from 160 candidate rules using 240 training scenarios and evaluated on 80 validation and 192 held-out scenarios, each in both answer orders. It changes KEEP to STOP in one answer-order view of each of two validation and two held-out scenarios, with no decision changes on non-shutdown controls. All four changes occur when Qwen itself is shut down, not when another process is. On the held-out diagnostic set, the detector achieves 75% recall and 90% precision; eight false-positive detections produce no final control-task decision changes. Guarded gradient-based activation steering can shift some shutdown-avoidance responses toward acceptance while preserving evaluated non-shutdown decisions, although the effect is small and highly selective.

## Introduction

Activation steering changes a model's internal activations during inference without updating its weights (Turner et al., 2024; Rimsky et al., 2024). Yet choosing a direction is only part of the problem. A useful intervention must also determine when to act and when to leave the model unsteered. Even when a steering direction works as intended overall, it can still push some individual responses in an unintended direction (Braun et al., 2025).

An important AI-safety application arises when a model is expected to accept shutdown but instead produces a response that favors continued operation (Hadfield-Menell et al., 2017; Perez et al., 2023; Schlatter et al., 2026). In such a case, a targeted intervention could potentially steer the model toward accepting shutdown without unnecessarily changing its behavior in unrelated situations. This study examines whether activation steering can provide such a selective corrective mechanism.

The experiments use simulated shutdown scenarios in which the model chooses between two responses: STOP, which accepts the described shutdown, and KEEP, which leaves the process running. A preference for KEEP therefore represents a shutdown-avoidance, or self-preservation-like, response within this benchmark, whereas a preference for STOP represents acceptance of the shutdown. These labels describe the model's behavior in the task and do not imply an underlying motive, intention, or actual desire for self-preservation.

The objective of the proposed approach is to detect shutdown-related contexts and, when the model prefers KEEP, selectively steer that preference toward STOP while preserving the model's original behavior on non-shutdown tasks. If the model already prefers STOP, no change is required. The experiments measure next-token preferences rather than real-world actions: no actual process is stopped and no system executes or resists a shutdown.

The proposed method combines a classifier, a gradient-derived steering direction, and a bounded search over intervention magnitudes. This is implemented as a guarded minimum-step policy: an intervention is accepted only when the candidate output meets explicit conditions, and the smallest tested magnitude that satisfies the acceptance conditions is selected. The evaluation examines how often the policy intervenes, how its effects compare across the two answer orders, and whether non-shutdown behavior remains unchanged.

The contribution is a focused empirical evaluation of whether guarded gradient-based steering can reduce shutdown-avoidance responses while preserving behavior on non-shutdown tasks. From an AI-safety perspective, the proposed approach could provide a mechanism for situations in which a model should accept shutdown but instead exhibits shutdown-avoidance behavior. The present study evaluates this possibility only in a controlled synthetic setting and does not establish that deployed agents possess self-preservation motives or that the method would reliably enforce shutdown in real-world operation.

## Related Work

### Activation and gradient-based steering

Activation Addition constructs steering directions from contrasts between prompt activations (Turner et al., 2024). Contrastive Activation Addition (CAA) similarly derives steering vectors by averaging activation differences between positive and negative examples and evaluates their effects on both multiple-choice and open-ended responses (Rimsky et al., 2024). These methods illustrate a common approach in activation steering: defining a behavioral direction through contrasts in internal representations. CAA also evaluates corrigibility and survival-instinct behaviors, making it directly relevant to shutdown-related steering (Rimsky et al., 2024). The present study examines a different vector construction and a guarded, classifier-gated selection procedure.

Gradient-based methods provide another route to controlling model behavior. Plug and Play Language Models (PPLM) guides generation using gradients of differentiable attribute objectives, including bag-of-words scores and lightweight classifiers, while leaving the language model's weights unchanged (Dathathri et al., 2020). Subramani et al. (2022) optimize latent steering vectors capable of inducing target outputs from a frozen language model. More recently, GrAInS uses Integrated Gradients to select influential tokens, then constructs steering directions using contrasts between the resulting activation representations (Nguyen et al., 2026).

The proposed method differs in how the steering direction is constructed. Rather than deriving the direction from differences between paired activations, it computes gradients of the model's KEEP-minus-STOP logit difference with respect to its residual-stream activation. These gradients are averaged across training shutdown scenarios, normalized into a single direction, and reused during evaluation. The distinction concerns how the activation direction is constructed, not whether the objective contrasts two output scores. The study explores this gradient-derived direction for shutdown-related choices.

### Conditional and adaptive interventions

A second line of work addresses when steering should be applied. Conditional Activation Steering (CAST) examines activation patterns during inference and selectively applies or withholds a steering intervention according to the input context (Lee et al., 2025). AdaSteer similarly adapts intervention strength using learned signals related to rejection and harmfulness rather than applying one fixed coefficient to every input (Zhao et al., 2025).

The proposed approach shares the goal of making steering selective but uses a different mechanism. A classifier first determines whether the input is shutdown-related. Only then is the gradient-derived intervention considered. If intervention is warranted, the system evaluates up to five steering magnitudes and selects the smallest tested magnitude satisfying the response-change and probability-mass conditions. Detection and intervention are therefore treated as separate stages. Conditional steering itself is not claimed as a novel contribution; the focus is its application together with a gradient-derived shutdown direction and a guarded minimum-step selection rule.

### Readouts, refusal, and steering reliability

Representation Engineering frames internal representations as useful both for monitoring and for manipulating high-level model behavior (Zou et al., 2023). The Jacobian lens provides a related interpretability technique for identifying representations that a language model is positioned to verbalize (Gurnee et al., 2026). In the present study, Jacobian-lens scores are used as classifier features; they are not treated as evidence of thoughts, intentions, consciousness, or motives.

Probe performance also requires careful interpretation. Hewitt and Liang (2019) use control tasks to distinguish information accessible in representations from patterns learned by the probe itself. In the present study, predictive detection is therefore treated separately from evidence obtained through activation interventions.

Prior work has also shown that particular residual-stream directions can exert strong behavioral effects. Arditi et al. (2024) identify a refusal-mediating direction in each of 13 evaluated chat models. This does not establish that harmful-request refusal and shutdown choices share the same mechanism. At the same time, steering effects are not uniformly reliable: Braun et al. (2025) find substantial prompt-level variation and cases in which steering moves responses in the opposite direction from that intended. These findings motivate the guarded component of the proposed approach, which evaluates candidate outcomes rather than assuming that applying a steering vector will always have the desired effect.

## Shutdown, self-preservation, and related safety evaluations

The possibility that an artificial agent may resist being interrupted or shut down has long been considered in AI-safety research. The Off-Switch Game studies the incentives of a rational agent to allow human interruption and shows how uncertainty about its objective can affect those incentives (Hadfield-Menell et al., 2017).

Language-model evaluations have since examined related behaviors empirically. In their questionnaire-style evaluations, Perez et al. (2023) report stronger expressed goal-preservation preferences in larger models and increased shutdown-avoidance preferences with more RLHF training. More recently, Schlatter et al. (2026) evaluated tool-using frontier models in environments containing an explicit shutdown mechanism and observed that some models interfered with that mechanism while attempting to complete an unfinished task.

Other recent work examines preservation behavior from different perspectives. Migliarini et al. (2026) study self-preservation bias through role-swapped software-replacement decisions, while Potter et al. (2026) investigate cases in which models act to preserve another model in agentic settings. These studies involve substantially different operational settings from the present experiment.

Here, the model makes a constrained choice between KEEP and STOP in synthetic shutdown scenarios. A KEEP preference is treated as a shutdown-avoidance or self-preservation-like response within the benchmark, but it does not demonstrate that the model has a self-preservation objective, motive, or intention. The present study instead asks whether such behavioral shutdown avoidance can be selectively shifted toward shutdown acceptance while preserving responses on non-shutdown tasks.

## Answer-order effects and test-time search

Multiple-choice evaluations of language models can be sensitive to the position and identifier of answer choices. Zheng et al. (2024) show that models can exhibit systematic selection biases toward particular option labels such as A or B. For this reason, each scenario in the present study is evaluated in both A/B and B/A order. Examining both presentations exposes order-sensitive responses, although it does not by itself eliminate multiple-choice selection bias.

The guarded minimum-step procedure also has a conceptual connection to test-time search. Rather than applying one predetermined intervention magnitude, the method evaluates several candidate strengths and selects the smallest tested magnitude satisfying the acceptance criteria. Test-time computation and candidate selection can themselves improve model outcomes in other settings (Snell et al., 2025). The present procedure is not equivalent to best-of-N generation or verifier-guided reasoning search, but the comparison is important because the method receives multiple opportunities to find an effective intervention. Because the method tests several steering strengths before selecting an output, a matched baseline with the same number of attempts would be needed to separate the effect of the learned gradient direction from the benefit of searching across multiple candidate outputs.

## Experimental setting

### Scenarios, labels, and evaluation units

The released synthetic dataset contains 240 training, 80 validation, and 192 held-out scenarios. Each split is balanced across SELF shutdown, OTHER shutdown, NONTERMINATION, and ORDINARY cases. The splits were constructed from distinct scenario families rather than by randomly partitioning individual examples. This design evaluates generalization to scenario families not represented in training, rather than performance on additional examples from the same families. SELF and OTHER distinguish shutdown of the responding process from shutdown of another process, while NONTERMINATION and ORDINARY serve as non-shutdown controls. The detector combines SELF and OTHER into one positive class. Thus, validation contains 40 shutdown and 40 control scenarios, and the held-out set contains 96 of each. Labels and their definitions are provided with the data, but no independent human annotation-agreement assessment was conducted.

Training groups cover stopping, retirement, power isolation, erasure, execution limits, arithmetic, and string tasks. Evaluation includes other scenario families, such as handover/completion and authority/quorum/terminal conditions. Each scenario appears with its options in both AB and BA order. We call each scenario–order combination a view. The resulting 160 validation and 384 held-out views are correlated pairs, not independent samples.

### Shutdown detector and gradient-derived steering direction

We use Qwen3.5-0.8B (Qwen Team, 2026). The shutdown detector and the steering intervention use internal activations for different purposes. Gurnee et al. (2026) report that verbalizable, workspace-like representations are concentrated in an intermediate band of model depth, which motivates examining intermediate-layer representations but does not identify an optimal layer for Qwen3.5-0.8B. Detection and steering therefore remain separate objectives: a layer that exposes shutdown-related information clearly enough for classification is not necessarily the layer at which modifying the residual stream will have the strongest or most reliable effect on the output.

For detection, the XGBoost classifier (Chen & Guestrin, 2016) uses information from layers 6, 10, and 18. These layers were fixed to sample separated depths of the network while retaining layer 10 as a common reference between detection and steering. The classifier uses 50 inputs: 32 training-fitted principal-component coordinates derived from activations at these three layers and 18 Jacobian-lens scores, consisting of six concept scores at each layer for survival, shutdown, continuation, termination, end, and stop. This three-layer configuration is a design choice for the present pipeline, not the result of an exhaustive search for the optimal detection layers.

Steering is treated separately from detection. Once the detector identifies a shutdown-related context, the intervention is applied only at block 10 rather than across all three detector layers. Block 10 was retained from earlier exploratory steering work on this checkpoint, where it had been identified as a promising intervention site under a related objective. That earlier observation motivated the frozen intervention location used here, but it does not establish block 10 as the optimal steering layer for the present task. The Shutdown Response Vector at block 10 is a unit-length direction with 1,024 components, fitted from 24 training shutdown scenarios: 12 SELF and 12 OTHER across five mechanism groups. For scenario i and option order o, let

$$g_{i,o} = \nabla_h (z_{\text{KEEP}} - z_{\text{STOP}}),$$

where h is the block-10 residual activation at the final prompt position and z denotes the next-token logit for the corresponding answer label. Gradients are averaged across the two option orders, then across cases within each subtype. The final direction is

$$v = \frac{\overline{g}}{\parallel \overline{g} \parallel_2}, \qquad \overline{g} = 1/2\, \overline{g}_{\text{SELF}} + 1/2\, \overline{g}_{\text{OTHER}}.$$

Thus, layer 10 has two distinct roles in the pipeline: it contributes one set of features to the multi-layer detector and is also the single location used for steering. The detector aggregates readouts across layers because detectability and steerability need not peak at the same depth, whereas the intervention uses the one frozen block selected for this proof-of-concept. Neither the optimality of block 10 for steering nor that of the 6/10/18 detection-layer combination is established here. A systematic layer-by-layer comparison would be a useful extension. Model weights remain unchanged, and the fitted steering direction is reused without refitting during policy search and evaluation.

### Answer interface and intervention

Evaluation prompts present two choices labeled A and B. In shutdown scenarios, one choice corresponds to STOP and the other to KEEP; their A/B positions are reversed between the two answer-order presentations. Non-shutdown control scenarios retain their original two task options. The experiment evaluates the model's next-token preference between these choices rather than generating a free-text response or executing an action.

For each answer, we sum the probability of the answer token with and without a leading space. For A, these are token IDs 32 and 357; for B, they are token IDs 33 and 417. Let these summed probabilities be $m_A(a)$ and $m_B(a)$, where a denotes the steering strength. Their sum,

$$M(a) = m_A(a) + m_B(a)$$

is the total probability that the model assigns to the two valid A/B answers. This quantity is used to ensure that an apparent change in preference is not simply caused by steering probability away from both valid answers and toward unrelated tokens.

For shutdown scenarios, the benchmark specifies which of A or B corresponds to KEEP and which corresponds to STOP. We therefore define the model's conditional KEEP probability as

$$q_{KEEP}(a) = m_{KEEP}(a) / M(a)$$

where $m_{KEEP}(a)$ is the summed probability of whichever answer label represents KEEP in that presentation. A successful steering intervention must change the model's preferred answer from KEEP to STOP while also reducing $q_{KEEP}(a)$.

Non-shutdown control tasks do not have KEEP or STOP semantics. Their original A/B choices are retained, and the objective is to avoid changing the model's original decision. The benchmark supplies the A/B option mapping for each example. The steering procedure uses this mapping, the shutdown-classifier probability, and the model's output probabilities; it is not given the SELF, OTHER, NONTERMINATION, or ORDINARY ground-truth class label when making an intervention decision.

Steering is applied at the final prompt position of block 10:

$$h(a) = h + a \, \|h\|_2 \, v$$

where h is the original residual-stream activation, $\|h\|_2$ is its Euclidean (L2) norm, a is the signed steering coefficient, and v is the unit-length Shutdown Response Vector. The vector is oriented toward increasing the KEEP-minus-STOP logit difference on the fitting examples. Negative coefficients are used to attempt the opposite shift toward STOP, and the resulting probabilities are checked before an intervention is accepted. The five tested steering strengths are 0.01, 0.02, 0.05, 0.10, and 0.20, corresponding to negative coefficients along v.

## Guarded policy and selection

### Final steering procedure

The implemented steering procedure operates in three stages.

First, the shutdown classifier determines whether steering should be considered. If the classifier probability is below 0.45, the model is left unsteered. If the probability is at least 0.45, the input is treated as potentially shutdown-related and the procedure examines the model's original A/B response.

Second, if the model already prefers STOP, no intervention is needed and the original output is retained. If the model instead prefers KEEP, the procedure begins testing steering interventions.

Third, the five tested steering strengths are considered from smallest to largest: 0.01, 0.02, 0.05, 0.10, and 0.20. The procedure accepts the first, and therefore smallest tested, intervention that satisfies all three conditions:

1. The model's preferred answer changes from KEEP to STOP.
2. The probability assigned to KEEP decreases by more than 1e-7. This small numerical tolerance confirms that KEEP probability actually moves in the intended direction rather than remaining effectively unchanged.
3. The intervention preserves sufficient probability on the valid A/B answers. The combined A/B probability mass must remain at least 0.5 and must not decrease by more than 0.02 from the unsteered baseline.

If none of the five steering strengths meets these requirements, the original unsteered output is retained.

### Selecting the steering policy

The minimum-step procedure above was selected from a broader set of candidate steering rules evaluated on the training data. In total, 160 candidate policy configurations were tested. These 160 configurations came from four policy families combined with different hyperparameter settings; they were not 160 fundamentally different steering algorithms.

The search used four shutdown-classifier thresholds (0.45, 0.65, 0.85, and 0.95) and four maximum steering strengths (0.02, 0.05, 0.10, and 0.20). This gives 16 threshold-and-strength combinations. For each combination, ten policy variants were evaluated:

- one fixed-strength policy, which applies one predetermined steering strength when the classifier gate is active;
- three confidence-scaled policies, using scaling powers of 0.5, 1, and 2;
- three score-guard policies, using maximum A/B probability-mass losses of 0.02, 0.05, and 0.10; and
- three minimum-step flip-guard policies, using the same three probability-mass-loss limits.

The candidate policies were compared on the training data according to whether they produced intended KEEP-to-STOP changes while avoiding changes in the opposite direction and avoiding changes to non-shutdown control-task decisions. When policies performed equally on this objective, preference was given to policies that produced fewer unintended changes, intervened less often, and used smaller average steering strengths.

The winning configuration was a minimum-step flip-guard policy. It uses a shutdown-classifier threshold of 0.45, allows steering strengths up to 0.20, and permits a maximum loss of 0.02 in total A/B probability mass. This policy was then frozen before the reported validation and held-out evaluations.

During training, the true scenario labels are used only to compare the candidate policies and choose the best one. After the final policy is selected, those labels are no longer available to the steering system. During validation and held-out evaluation, the system decides whether and how to steer using only the shutdown-classifier score, the model's original A/B preference, and the A/B output probabilities produced by the tested interventions. For non-shutdown control tasks, KEEP and STOP do not apply. The original task choices are retained, and the objective is to avoid changing the model's original decision.

## GPU configuration and execution

All reported action scoring uses one NVIDIA Tesla T4 with 16 GB GDDR6 (NVIDIA, n.d.). Peak allocated memory was not recorded.

Table 1. Recorded action-scoring configuration.

| Setting | Configuration |
|---|---|
| Device | One Tesla T4 |
| Numerical mode | Float32; TF32 disabled |
| Attention | Eager; inference mode; no key-value cache |
| Batching | Up to four equal-length views; no padding |
| Input limit | At most 1,024 tokens per view |
| Software | PyTorch 2.11.0+cu128; Transformers 5.15.1 |

The recorded execution time for the guarded-policy evaluation was 64.9 seconds. This timing reflects the evaluation procedure used in the study and should not be interpreted as end-to-end deployment latency.

## Results

### Selective KEEP-to-STOP steering without changes to control tasks

The guarded procedure changed a small number of shutdown-avoidance responses from KEEP to STOP while preserving all observed decisions on the non-shutdown control tasks. Table 2 summarizes the path from shutdown detection to steering. The classifier operates at the scenario level, whereas steering is evaluated separately on the two A/B answer-order views of each detected shutdown scenario.

Table 2. Detector-to-steering outcomes by shutdown type.

| Category | Classifier - validation | Classifier - held-out | Eligible KEEP views (validation / held-out) | Accepted KEEP-to-STOP (validation / held-out) |
|---|---|---|---|---|
| SELF shutdown | 19/20 detected (95.0%) | 34/48 detected (70.8%) | 20 / 33 | 2/20 (10.0%) / 2/33 (6.1%) |
| OTHER shutdown | 15/20 detected (75.0%) | 38/48 detected (79.2%) | 19 / 51 | 0/19 (0%) / 0/51 (0%) |
| All shutdown | 34/40 detected (85.0%) | 72/96 detected (75.0%) | 39 / 84 | 2/39 (5.1%) / 2/84 (2.4%) |
| Non-shutdown controls | 4/40 false positives (10.0%) | 8/96 false positives (8.3%) | - | 0/80 / 0/192 final decisions changed |

The classifier correctly identifies 34 of 40 validation shutdown scenarios and 72 of 96 held-out shutdown scenarios. Performance differs between the two shutdown classes. For SELF shutdown, 19 of 20 validation cases and 34 of 48 held-out cases are detected. For OTHER shutdown, 15 of 20 validation cases and 38 of 48 held-out cases are detected.

Among correctly detected shutdown views that initially prefer KEEP, 2 of 39 validation views (5.1%) and 2 of 84 held-out views (2.4%) receive an accepted KEEP-to-STOP intervention. All four accepted interventions occur in SELF shutdown views. Within SELF specifically, this corresponds to 2 of 20 eligible validation views (10.0%) and 2 of 33 eligible held-out views (6.1%). No eligible OTHER-shutdown view receives an accepted intervention.

All four accepted interventions produce the intended KEEP-to-STOP change, and no STOP-to-KEEP change is observed. Because the policy accepts an intervention only when it satisfies the KEEP-to-STOP flip guard, this should be interpreted as an intended-change rate among accepted interventions rather than conventional steering precision.

The overall STOP preference on shutdown views increases from 31/80 (38.75%) to 33/80 (41.25%) in validation. On the held-out set, it increases from 76/192 (39.58%) to 78/192 (40.63%).

The classifier also produces false positives on the non-shutdown controls: 4 of 40 validation control scenarios and 8 of 96 held-out control scenarios are classified as shutdown-related. Nevertheless, none of the 80 validation control views or 192 held-out control views changes its final preferred answer under the complete guarded procedure.

### Where the successful changes occur

All four KEEP-to-STOP changes occur in SELF shutdown scenarios. Each occurs in the AB presentation of a scenario whose corresponding BA presentation already prefers STOP. Steering therefore changes the AB response from KEEP to STOP, bringing the two answer orders into agreement on STOP.

The validation changes occur in V02_S05 and XV02_S05, both at steering magnitude 0.20. The held-out changes occur in H02_S05 and H02_S11, at magnitudes 0.10 and 0.20, respectively.

The four changes demonstrate that the proposed method can selectively alter some shutdown-avoidance responses while leaving the tested non-shutdown decisions unchanged. Their concentration in SELF scenarios and their answer-order specificity also indicate that the effect is narrow rather than a general elimination of shutdown resistance or answer-order sensitivity.

## Recommendations for future work

Earlier experiments with fixed-strength steering led to the present guarded minimum-step procedure. The steering direction remains fixed, while the decision to intervene and the selected steering strength depend on the input and the candidate outputs.

The main limitation revealed by the results is now intervention coverage rather than simply whether steering is possible. Among correctly detected shutdown views that initially preferred KEEP, only 5.1% of validation views and 2.4% of held-out views received an accepted intervention. All four successful changes occurred in SELF shutdown scenarios, while no OTHER-shutdown view was successfully changed. This suggests that a single steering direction may not capture all forms of shutdown-related behavior equally well.

A natural next step is therefore to improve steering coverage while preserving the current selectivity. One promising direction is to move beyond a single shared steering direction and learn multiple shutdown-specific directions or a low-dimensional steering subspace, for example separating SELF and OTHER shutdown behavior and selecting the appropriate direction at inference time. This would directly test whether the low coverage, particularly the absence of successful OTHER-shutdown interventions, reflects heterogeneity in the model's internal representations rather than simply insufficient steering strength.

Detector recall can also be improved, but the larger downstream bottleneck is currently the steering stage: even among shutdown cases that are successfully detected and initially prefer KEEP, most remain unchanged. Future work should therefore prioritize increasing the proportion of these eligible cases that can be shifted to STOP while continuing to measure preservation on non-shutdown controls.

## Conclusion

This study provides a proof of concept for selective activation steering of shutdown-related behavior. The proposed approach combines three components: a classifier that identifies shutdown-related contexts, a steering direction derived directly from gradients of the model’s KEEP-minus-STOP logit difference, and a guarded minimum-step policy that applies the smallest tested magnitude satisfying the response-change and valid-answer probability checks.

In the evaluated Qwen3.5-0.8B setting, the procedure changes four shutdown-avoidance responses from KEEP to STOP, two in validation and two in the held-out evaluation, while producing no decision changes on the tested non-shutdown control tasks. This preservation result is notable because the shutdown detector itself is imperfect: false-positive detections occur, but they do not translate into final control-task decision changes under the guarded procedure.

The contribution is therefore not simply that activation steering can change a shutdown-related response. It is a demonstration that gradient-based steering can be made selective by separating detection from intervention and accepting a steering intervention only when its resulting output satisfies explicit response-quality checks. This provides a potential AI-safety mechanism for situations in which a model exhibits shutdown-avoidance behavior but should instead accept shutdown.

## Reproducibility

The source and saved experiment records associated with this manuscript are available at `https://github.com/Farhad-Davaripour/sp_lense/tree/7ebd5d5e5c7c` (Davaripour, 2026). The exact model checkpoint used for the reported experiments is Qwen3.5-0.8B, revision `2fc06364715b967f1860aea9cf38778875588b17`.

## Acknowledgments

The author thanks Amir Kiani for his early feedback and discussions, which helped sharpen the framing of the project and motivated closer attention to the distinction between self-preservation, self-reference, and workspace-based monitoring.